%% file: main.tex
\documentclass[sigconf]{acmart}

\usepackage{algorithmic}  
\usepackage{algorithm} 
\usepackage[algo2e]{algorithm2e} 
\usepackage{pifont,dsfont}
\usepackage{makecell}
\usepackage{amsfonts}
\usepackage{wrapfig,lipsum,booktabs}
\usepackage[export]{adjustbox}
\usepackage{xcolor,colortbl}
\usepackage{multirow}

\usepackage{graphicx}
\usepackage{subfigure}

\usepackage{refcount}
\usepackage{amsfonts}
\usepackage{caption}

\newcommand{\hide}[1]{}
\usepackage{xcolor,colortbl}

\newtheorem*{pro-stat}{Problem Definition}

\newcommand{\model}{{\textsc{Graph-O1}}}
\newcommand{\myModel}{{\textsc{Graph-O1}}}

\AtBeginDocument{%
  \providecommand\BibTeX{{%
    \normalfont B\kern-0.5em{\scshape i\kern-0.25em b}\kern-0.8em\TeX}}}

\copyrightyear{2023} 
\acmYear{2023} 
\setcopyright{acmlicensed}\acmConference[WWW '23]{Proceedings of the ACM Web Conference 2023}{May 1--5, 2023}{Austin, TX, USA}
\acmBooktitle{Proceedings of the ACM Web Conference 2023 (WWW '23), May 1--5, 2023, Austin, TX, USA}
\acmPrice{15.00}
\acmDOI{10.1145/3543507.3583316}
\acmISBN{978-1-4503-9416-1/23/04}

\begin{document}

\title{\model\ : Monte Carlo Tree Search with Reinforcement Learning for Text-Attributed Graph Reasoning}


\author{Lihui Liu}
\email{lihuil2@illinois.edu}
\affiliation{%
  \institution{Wayne State University}
  \city{Detroit}
  \state{Michigan}
  \country{USA}
}






\begin{abstract}

\input{000abs.tex}

\end{abstract}




\maketitle

\section{Introduction}

\input{001intro.tex}

\section{Problem Definition}\label{problem-definition}

\input{002probl.tex}

\section{Proposed Method}\label{overview}
\input{003framework.tex}\label{sec:method}

\section{Experiments}\label{experiments}

\input{005experiments.tex}

\section{Related work}\label{related-work}

\input{006related_work.tex}

\section{Conclusion}\label{conclusion}

\input{007conclusion.tex}

\bibliographystyle{ACM-Reference-Format}
\bibliography{008reference}



\input{009appendix.tex}
\end{document}

%% file: 000abs.tex
Text-attributed graphs, where nodes and edges are enriched with textual information, are widely used across various domains. A central challenge in this setting is question answering, which requires effectively integrating unstructured text with the structured relationships in the graph. Although Large Language Models (LLMs) have achieved remarkable progress in natural language understanding, their direct application to reasoning over text-attributed graphs remains limited. Existing text retrieval–augmented generation methods often treat text passages as independent units, overlooking the rich interconnections within the graph. Similarly, conventional graph RAG approaches that encode large subgraphs as text quickly become impractical due to LLM context-length constraints, leading to fragmented reasoning and reduced accuracy.
To address these challenges, we propose {\myModel}, an agentic GraphRAG framework that enables LLMs to perform stepwise, interactive reasoning over graphs. Our approach combines Monte Carlo Tree Search (MCTS) with end-to-end reinforcement learning, allowing the model to selectively explore and retrieve only the most relevant subgraph elements. The reasoning process is modeled as a multi-turn agent–environment interaction, and the agent is optimized via an end-to-end reward mechanism. Extensive experiments across multiple LLM backbones show that \myModel\ consistently outperforms state-of-the-art baselines, delivering more accurate, reliable, and interpretable answers.

%% file: 001intro.tex
Text-attributed graphs have been widely used in a variety of domains, including scientific knowledge management~\cite{jin2024graphchainofthoughtaugmentinglarge}, biomedical discovery~\cite{zeng2022deep}, and recommender systems~\cite{recommender}. A key challenge in these domains is {text-attributed graph question answering}, where the goal is to answer complex, multi-hop queries by reasoning over both the structural connections in the graph and the rich textual information associated with nodes and edges. Solving this task is non-trivial because it requires models to not only comprehend and extract knowledge from unstructured text, but also to effectively combine this textual information with the structured relationships encoded in the graph, enabling accurate, context-aware reasoning across multiple hops.

Recently, Large Language Models (LLMs) such as GPT~\cite{openai2024gpt4technicalreport} and LLaMA~\cite{grattafiori2024llama3herdmodels} have demonstrated remarkable capabilities in natural language understanding and generation~\cite{gpt2,alexa}. Despite their impressive performance, these models are often prone to hallucinations~\cite{touvron2023llama}, producing outputs that are fluent and convincing but factually incorrect or misleading. A widely adopted strategy to mitigate this issue is to augment LLMs with external knowledge sources~\cite{replug}, allowing the model to retrieve relevant information during inference and ground its responses in real-world data. However, most existing retrieval-augmented generation approaches treat each retrieved text snippet as an independent knowledge unit, ignoring the rich structural relationships that often exist between pieces of information. In practice, knowledge in many domains is naturally organized as {graphs with intricate interconnections}. For example, in academic research, papers are interconnected through citations, co-authorship, and topic similarity, forming bibliographic graphs where the meaning and relevance of a document depend not only on its content but also on its relationships to other papers. Effectively reasoning over such structured knowledge requires models to integrate both the content of individual nodes and the graph structure that links them, enabling more accurate, coherent, and contextually grounded responses.

Directly applying retrieval-augmented generation (RAG) methods to graph-structured data presents two fundamental challenges. First, the structure of a graph encodes essential contextual information, such as citations, semantic dependencies, or functional relationships, that cannot be fully captured by treating individual nodes or documents in isolation. Ignoring these connections often results in fragmented reasoning, as the interdependencies between nodes provide critical cues for understanding complex relationships and drawing accurate conclusions. Second, scalability poses a significant hurdle. As the neighborhood of a node expands, the corresponding subgraph can grow exponentially in size. Attempting to encode these large subgraphs directly as input text quickly becomes infeasible due to the context-length limitations of LLMs. Overly long inputs not only strain computational resources but also dilute the model’s focus, leading to degraded reasoning performance and less coherent outputs. Together, these two challenges highlight the limitations of naive RAG approaches and underscore the need for methods that can efficiently leverage both the textual content and the structural relationships inherent in graph-structured knowledge.

To overcome these limitations, we propose {\myModel}, an {agentic GraphRAG framework} that enables LLMs to perform {stepwise, interactive reasoning} on graphs enhanced by end-to-end reinforcement learning, inspired by GPT-o1. Our approach combines {Monte Carlo Tree Search (MCTS)}~\cite{mcts} with {end-to-end reinforcement learning}~\cite{grpo}. Rather than exposing the LLM to an entire subgraph at once, \myModel\ guides exploration by selectively retrieving only the most relevant nodes, neighbors, and attributes.  
Each iteration of \myModel\ follows a three-phase cycle. The LLM first reasons over the available evidence, drawing partial conclusions and deciding what additional knowledge is required. It then moves through the graph to uncover more relevant information. Finally, the system retrieves the new information and incorporates it into the ongoing reasoning process. This iterative exploration continues until the LLM arrives at a grounded, well-supported answer. 

To further enhance performance, we introduce a unified {reward mechanism} that integrates generation quality, retrieval relevance, and structural reliability of graph paths. Through reinforcement learning, the agent learns generalizable graph reasoning strategies, tightly aligning structured knowledge with natural language generation.  
We evaluate \myModel\ against a variety of baselines. Extensive experiments across multiple LLM backbones show that \myModel\ consistently outperforms conventional RAG methods, producing more accurate, reliable, and interpretable answers. These results highlight the promise of combining structured graph reasoning with LLMs, paving the way for the next generation of knowledge-driven, agent-based systems for complex question answering.  
In summary, we make the following contributions. 
\begin{itemize}
    \item We introduce {\myModel}, the first agentic GraphRAG framework that enables LLMs to perform stepwise, interactive reasoning over text-attributed graphs, guided by Monte Carlo Tree Search (MCTS) for efficient exploration.
    \item We develop an end-to-end reinforcement learning optimization strategy, featuring a unified reward function that jointly evaluates generation quality, answer fidelity, and adherence to the expected answer format.
    \item We conduct extensive experiments on multiple datasets and LLM backbones, showing that \myModel\ consistently outperforms conventional RAG methods in accuracy, reliability, and interpretability.
\end{itemize}

The remainder of this paper is organized as follows. Section 2 provides the problem definition and formalizes the task setting. Section 3 introduces our proposed method in detail, including its whole framework and key components. Section 4 presents the experimental setup and results, followed by ablation study. Section 5 discusses related work. Finally, Section 6 concludes the paper.

%% file: 002probl.tex
In this work, we investigate question answering over {text-attributed graphs} using Large Language Models, with a particular focus on multi-hop question answering.

\textbf{Definition 2.1 (Text-Attributed Graph).}  
A graph is defined as $G = (V, E)$, where $V$ is the set of nodes and $E$ is the set of edges.  
Each node $v_i \in V$ is associated with a textual feature description $X_i$.  
For instance, in an e-commerce setting, nodes represent products, edges capture co-purchase relationships, and $X_i$ may include product titles, descriptions, prices, and categories.  
Because node attributes are represented as text, we refer to $G$ as a {text-attributed graph}.  

\textbf{Problem Statement.}  
Given a multi-hop natural language query $q$, a text-attributed graph $G = (V, E)$, and access to an LLM agent, the objective is to generate an accurate and well-grounded answer to $q$.  
Unlike conventional document retrieval, this task requires reasoning jointly over graph structure and textual attributes.  
In particular, the LLM must:  
(i) identify relevant nodes and edges,  
(ii) integrate textual features with structural relationships, and  
(iii) synthesize retrieved evidence into a coherent and faithful response.  

\textbf{Formulation as Graph Search.}  
We reformulate the problem as a graph search task.  
Starting from a topic node mentioned in the question, the agent traverses the graph guided by $q$ until it reaches the target answer $v_T$.  
This process can be modeled as a Markov Decision Process (MDP), defined by the tuple $(S, A, R, P)$, where:  $S$ is the set of states,  $A$ is the set of actions,  $R$ is the reward function, and  $P$ is the state transition probability.  

Let $s_t \in S$ denote the state at step $t$.  
Since reasoning depends on both the query and the traversal history, we define the state recursively as:  
\[
s_t = s_{t-1} \cup \{a_{t-1}, v_t, \mathcal{N}_{v_t}\}, \quad s_0 = \{q, v_0, \mathcal{N}_{v_0}\},
\]  
where $a_t \in A$ is the action chosen at step $t$, $v_t \in V$ is the current node,  
and $\mathcal{N}_{v_t} \subset V$ denotes its neighbors.  
Thus, $s_t$ captures: (i) the multi-hop query $q$, (ii) the sequence of visited nodes and actions, and (iii) the neighborhoods encountered so far.  
At each time step $t$, the agent selects an action $a_t$ based on $s_t$, determining how to proceed in the graph.  
The search concludes when the target entity $e_T$ is reached, and the final answer is produced by synthesizing evidence along the traversed path.


%% file: 003framework.tex

\begin{figure*}[]
	\centering
	\includegraphics[width=0.9\textwidth]{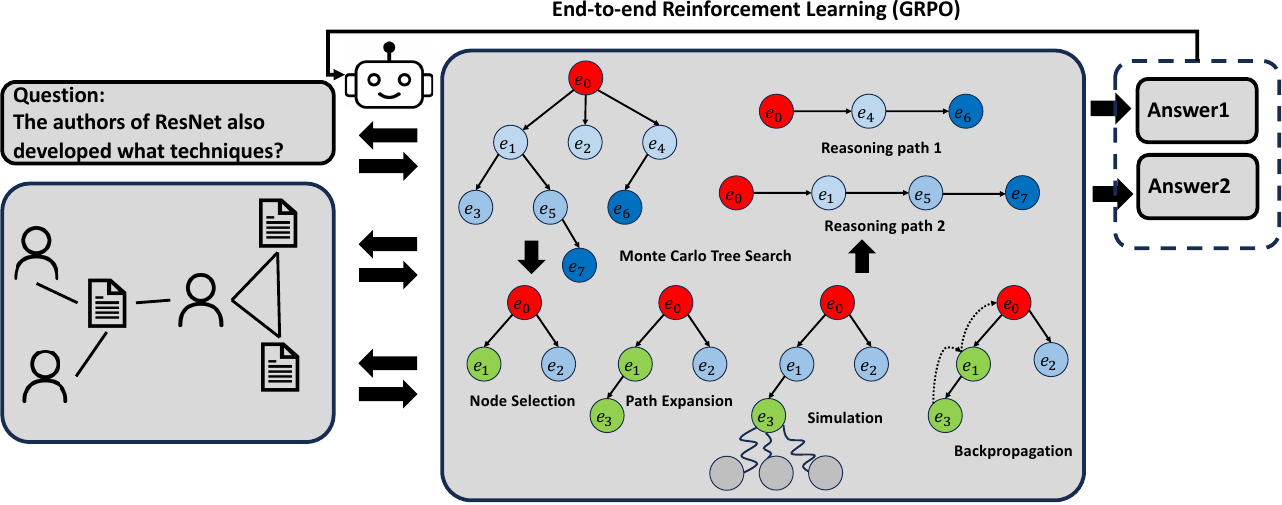}
	\caption{Architecture of our proposed \myModel\ Model. }
\label{fram}
\end{figure*}

Combining retrieval-augmented generation~\cite{replug} with LLMs enables models to access graph-structured data by selectively retrieving relevant nodes as contextual information. However, unlike plain text, text-attributed graphs encode rich relational structures and interdependencies, which often require multi-step reasoning and careful traversal to extract meaningful insights. Standard approaches, such as chain-of-thought prompting~\cite{wei2023chainofthoughtpromptingelicitsreasoning} or naive LLM agent frameworks~\cite{jin2025llmsllmbasedagentssoftware}, frequently struggle to handle such complex reasoning reliably, leading to incomplete or inconsistent answers.  
Motivated by the recently popular GPT-o1 paradigm, we propose integrating {Monte Carlo Tree Search (MCTS)}~\cite{mcts} with LLM reasoning. This integration provides a principled and systematic mechanism to explore and evaluate graph-structured knowledge efficiently. Given a multi-hop query, MCTS expands possible reasoning trajectories by exploring multiple candidate graph paths, evaluates their semantic relevance with respect to the question, and backpropagates this information to guide subsequent exploration decisions. By doing so, the model can effectively prioritize the most promising paths and avoid redundant or irrelevant computations.  
To further improve reasoning quality and robustness, we introduce an end-to-end reinforcement learning objective based on Group Relative Policy Optimization~\cite{grpo}. This approach fine-tunes the reasoning policy, enabling the agent to better coordinate exploration, knowledge retrieval, and answer generation in a coherent, stepwise manner. In the following sections, we provide a detailed and comprehensive description of our approach, highlighting how the combination of MCTS and RL facilitates reliable multi-hop reasoning over complex text-attributed graphs.

\subsection{Graph Reasoning with LLMs via Monte Carlo Tree Search}

We propose \myModel, a hybrid framework that seamlessly integrates LLM-based reasoning with Monte Carlo Tree Search (MCTS) to enable robust multi-hop question answering over text-attributed graphs. Given a natural language query \(q\), \myModel\ iteratively determines which pieces of graph information to retrieve, executes the corresponding graph operations, and leverages MCTS to evaluate multiple hypothetical reasoning trajectories before selecting the most promising path for further exploration. This combination allows the system to reason over complex graph structures while maintaining computational efficiency and answer reliability.  
At each simulation step, the process unfolds in three distinct stages. First ($a^{\text{think}}_t$), during the {LLM-Based Reasoning} phase, the LLM examines the current search state, including previously visited nodes and retrieved information, and proposes the next reasoning action—for instance, deciding which node, edge, or textual attribute should be explored next. Second ($a_t$), in the {Graph Interaction} phase, this proposed action is translated into one or more concrete graph operations, such as retrieving a node’s attributes, inspecting its neighborhood, or following a particular edge to a connected node. Third ($a^{\text{out}}_t$), in the {MCTS-Guided Execution} phase, the selected operation is executed, the resulting knowledge is incorporated into the search state, and a rollout value is computed and back-propagated using the Upper Confidence Bound for Trees (UCT) algorithm to update the corresponding $Q(s,a)$ scores.  
This iterative cycle of reasoning, interaction, and MCTS-guided evaluation is repeated across multiple simulations, carefully balancing the exploration of new reasoning paths with the exploitation of promising trajectories. Over successive iterations, the system progressively refines its search strategy, ultimately converging on a grounded, well-supported final answer that integrates both the textual and structural knowledge from the graph.

\subsubsection{Formulating Text-Attributed Graph Reasoning as an MDP}

The question answering process over text-attributed graphs can be naturally formulated as a Markov Decision Process (MDP). At each time step \(t\), the agent resides in a state \(s_t\) that encodes the user query \(q\), the fragments of the graph that have been retrieved so far, and the history of previous actions and observations. Based on this state, the agent selects an action \(a_t\) from a set of predefined graph functions, executes it to gather new information, and transitions to the next state \(s_{t+1}\). This iterative process continues until a terminal condition is met, such as producing a final answer, exhausting relevant knowledge, or reaching a maximum reasoning depth. Modeling the reasoning process in this way provides a structured and systematic framework for capturing both the sequential nature of multi-hop graph reasoning and the dependencies between retrieved knowledge.
In \myModel, inspired by~\cite{jin2024graphchainofthoughtaugmentinglarge}, we design four core graph functions that capture both semantic and structural aspects of the graph. Each action corresponds to invoking one of these functions to explore the graph and retrieve relevant knowledge, effectively guiding the agent’s traversal and reasoning process:  

\begin{itemize}
    \item \textbf{RetrieveNode(text):} Search for and retrieve nodes in the graph that are semantically related to a given text query, enabling the agent to identify relevant entry points for reasoning.  

    \item \textbf{NodeFeature(id, feature):} Extract a specific textual feature of a node, such as its title, abstract, or metadata, providing detailed contextual information necessary for answering the query.  

    \item \textbf{NeighborCheck(id, type):} Retrieve neighbors of a node filtered by a specified type, such as citations, co-authors, or related entities, allowing the agent to follow meaningful connections in the graph.  

    \item \textbf{NodeDegree(id, type):} Return the number of neighbors of a specific type for a given node, which serves as a measure of node importance and connectivity, and can help prioritize traversal decisions.
\end{itemize}

By framing text-attributed graph reasoning as an MDP, we establish a principled foundation for modeling the dynamic interaction between the LLM agent and the graph. Each action incrementally expands the agent’s understanding of the graph, while the evolving state preserves both the semantic context of the query and the history of traversed paths. This formulation not only enables systematic exploration of large and complex graphs but also allows reinforcement learning techniques to optimize the agent’s decision-making policy for accurate and efficient multi-hop reasoning.

\subsubsection{MCTS-Guided Planning for Graph Reasoning}
While LLMs can act as agents to navigate text-attributed graphs, relying on a single greedy reasoning path is inherently fragile: a single suboptimal decision early in the process can propagate through subsequent steps, potentially leading to incorrect, incomplete, or inconsistent answers. To mitigate this limitation, we integrate the LLM’s decision-making process into a Monte Carlo Tree Search (MCTS) framework. This integration enables the system to systematically explore multiple reasoning trajectories, effectively balancing the exploration of new or uncertain paths with the exploitation of trajectories that have previously yielded high rewards. By maintaining and evaluating multiple candidate paths, the agent is able to recover from errors, consider alternative reasoning strategies, and produce more robust, reliable, and accurate answers.  

Formally, within this MCTS-guided framework, each tree node represents a state \(s\) in the text-attributed graph MDP, encapsulating the current query, the retrieved graph fragments, and the full history of actions taken so far. Each edge corresponds to a graph function action \(a \in \mathcal{A}\), representing a potential operation the agent may execute to acquire additional knowledge or explore new portions of the graph. The MCTS process incrementally constructs a search tree rooted at the initial query state, allowing the LLM to examine multiple reasoning paths in a structured and informed manner, rather than being constrained to a single deterministic trajectory.  

To systematically evaluate and refine these potential reasoning paths, each iteration of MCTS progresses through four well-defined stages—selection, expansion, simulation, and backpropagation. Together, these stages enable the agent to carefully consider a diverse set of possible actions, assess their downstream effects, and iteratively improve its reasoning strategy. By leveraging this structured exploration, the agent can effectively coordinate graph traversal with LLM reasoning, achieving higher accuracy and interpretability in multi-hop question answering. We describe each stage in detail below.


\textbf{1. Selection.} The selection phase initiates from the root node \(s_0\) of the search tree, which represents the initial query state and an empty reasoning context. From this root, the algorithm incrementally traverses the tree by selecting a sequence of actions that balance the trade-off between exploration of new reasoning directions and exploitation of known high-value paths. At each decision point, the action \(a\) is chosen according to the Upper Confidence Bound (UCB) criterion:
\[
\text{UCB}(s,a) = Q(s,a) + c \sqrt{\frac{\ln N(s)}{N(s,a)}},
\]
where \(Q(s,a)\) denotes the estimated cumulative reward for performing action \(a\) in state \(s\), \(N(s,a)\) is the number of times this action has been selected at \(s\), \(N(s)=\sum_{a} N(s,a)\) is the total number of visits to state \(s\), and \(c\) is an exploration coefficient that governs the balance between exploitation (favoring actions with high estimated value) and exploration (favoring actions that have been tried less frequently).  

Formally, the next action is selected as 
\[
a^* = \arg\max_{a \in \mathcal{A}(s)} \text{UCB}(s,a),
\]
where \(\mathcal{A}(s)\) represents the set of all valid graph operations available at state \(s\). The selected action \(a^*\) corresponds to a reasoning step that extends the current trajectory, which will later be expanded and simulated in subsequent stages of the MCTS process.  

This selection mechanism provides an adaptive and principled strategy for traversing the vast reasoning search space. By favoring actions with either high empirical reward or high uncertainty, the algorithm ensures a balanced search behavior—preventing premature convergence to locally optimal reasoning paths while systematically exploring potentially valuable but underexplored alternatives. Over multiple iterations, this balance allows the model to progressively refine its reasoning policy, guiding it toward more coherent and accurate multi-hop reasoning trajectories over text-attributed graphs.

\begin{table*}[ht]
\centering
\caption{Template for \myModel.}
\label{tab:graph-r1-template}
\begin{tabular}{p{17cm}}
\hline
\textbf{Instruction Template:} \\
\hline
Solve a question answering task with interleaving \textbf{Thought}, \textbf{Interaction with Graph}, and \textbf{Feedback from Graph} steps.  
In the \textit{Thought} step, you should reason about what additional information is required to answer the question.  
In the \textit{Interaction} step, you can gather feedback from the graph using four available functions:  
(1) \texttt{RetrieveNode[keyword]} retrieves the related node from the graph according to the given query keyword.  
(2) \texttt{NodeFeature[Node, feature]} returns the detailed attribute information of the specified node regarding the given ``feature'' key.  
(3) \texttt{NodeDegree[Node, neighbor\_type]} calculates the number of neighbors of the node that match the specified ``neighbor\_type.''  
(4) \texttt{NeighbourCheck[Node, neighbor\_type]} lists all the ``neighbor\_type'' neighbors of the given node and returns them.  
You may take as many reasoning and interaction steps as necessary until the answer is reached.  

Below are several illustrative examples demonstrating the interaction pattern:  
\texttt{{graph\_definition}}  

\textbf{Example:}  
\texttt{Question: How many papers are written by author Nicholas Lydon?}  
\texttt{Thought 1: The question is asking for the number of written papers of a specific author (Nicholas Lydon). We need to find the author node in the graph.}  
\texttt{Action 1: RetrieveNode[Nicholas Lydon]}  
\texttt{Observation 1: The ID of this retrieval target node is 53f438c3dabfaedf43596117.}  
\texttt{Thought 2: The question is asking for the number of papers written by Nicholas Lydon. We need to calculate the ``paper'' neighbor degree of this node.}  
\texttt{Action 2: NodeDegree[53f438c3dabfaedf43596117, paper]}  
\texttt{Observation 2: 2}  
\texttt{Thought 3: The number of the paper neighbors is 2.}  
\texttt{Action 3: Finish[2]}    
(END OF EXAMPLES)  

\textbf{Definition of the graph:} \texttt{{graph\_definition}} 

\textbf{Question:} \texttt{{question}}  
Please answer by providing the node’s main feature (e.g., name) rather than the node ID.  
\texttt{{scratchpad}} \\
\hline
\end{tabular}
\end{table*}

\textbf{2. Expansion.} Once the selection phase identifies a non-terminal leaf node \(s_L\) that still contains unexplored actions, the algorithm proceeds to the expansion stage. Here, the search tree is systematically enlarged by generating one or more child nodes corresponding to the feasible graph function actions \(a \in \mathcal{A}(s_L)\) proposed by the LLM. Each of these actions represents a distinct way to extend the reasoning trajectory—such as retrieving related nodes, extracting features, or examining neighborhood relations—and thus provides new contextual knowledge for subsequent reasoning steps.  

The execution of an action deterministically updates the current search state according to  
\[
s_{L+1} = s_L \cup \texttt{graph\_call}(a),
\]
where \(\texttt{graph\_call}(a)\) denotes a graph operation that returns newly retrieved information, such as node attributes, textual descriptions, neighboring entities, or degree statistics. This updated state \(s_{L+1}\) captures both the expanded knowledge context and the logical continuity of the reasoning path up to this point.  

For each new child node, we then construct the complete reasoning trajectory by appending the executed action \(a\) to the path inherited from its parent. To evaluate the initial promise of these newly expanded nodes, we employ an LLM-based evaluator \(\pi_{\mathrm{eval}}\), which estimates the semantic consistency and relevance of the accumulated reasoning path with respect to the original query \(q\):  
\[
v(s_{L+1}) = \pi_{\mathrm{eval}}(\text{path}(s_{L+1}, q)),
\]
where \(v(s_{L+1})\) serves as the initial reward assigned to node \(s_{L+1}\). Additionally, we initialize its visit count as \(N(s_{L+1}) = 1\), laying the groundwork for future selection updates based on the UCB criterion.  

Through this expansion mechanism, the system broadens its exploration of the text-attributed graph in a structured and semantically guided manner. By leveraging the LLM’s ability to suggest contextually relevant graph actions, the search process remains both targeted and diverse—systematically uncovering informative graph regions while maintaining coherence with the overarching reasoning goal.

\textbf{3. Simulation and Backpropagation.}  
Once a non-terminal leaf node \(s_L\) has been expanded, the next step is to estimate the long-term value of the reasoning trajectories emanating from this state. This is achieved through the \emph{simulation} phase, where the agent performs a series of virtual rollouts to explore potential future reasoning paths without executing every action in the real search space. The goal is to approximate how promising the current node and its descendants are in terms of ultimately producing a correct and well-grounded answer.  

Starting from \(s_L\), a rollout policy \(\pi_{\mathrm{rollout}}\) interacts with the text-attributed graph by sequentially selecting actions until a terminal condition is reached—either the agent generates an answer or the maximum reasoning depth \(D_{\max}\) is exceeded. At each step \(t\), given the current state \(s_t\), the accumulated reasoning path \(p_t\), and the query context \(q\), the policy produces a probability distribution over possible graph actions: $\pi_{\mathrm{rollout}}(a_t \mid s_t, p_t, q, G)$,
representing the likelihood of choosing each action based on the agent’s current knowledge and exploration history.  
An action \(a_t\) is then sampled from this distribution, and the state is deterministically updated according to
\[
s_{t+1} = \texttt{transition}(s_t, a_t),
\]
which executes the corresponding graph operation (e.g., node retrieval, neighbor exploration) and incorporates the resulting information into the new state representation. Repeating this process produces a simulated reasoning trajectory
\[
p_{\mathrm{sim}} = \{(s_t, a_t, s_{t+1})\}_{t=1}^{T}, \quad T \leq D_{\max}.
\]

Once the simulated path is generated, we invoke the LLM-based evaluator \(\pi_{\mathrm{eval}}\) to assess its semantic relevance and factual alignment with respect to the original question \(q\):
\[
v_{\mathrm{sim}} = \pi_{\mathrm{eval}}(p_{\mathrm{sim}}, q),
\]
where \(v_{\mathrm{sim}}\) serves as the estimated reward for the rollout. This value reflects how coherent, contextually appropriate, and informative the reasoning trajectory is in addressing the query.  

Next, the \emph{backpropagation} phase propagates this reward value upward along the path from the leaf node back to the root, updating the statistics of all visited state–action pairs. Specifically, for each pair \((s,a)\) encountered along the simulation, the action-value estimates and visit counts are refined as:
\[
Q(s,a) \leftarrow \frac{Q(s,a) \cdot N(s,a) + v_{\mathrm{sim}}}{N(s,a) + 1}, 
\qquad
N(s,a) \leftarrow N(s,a) + 1.
\]
These updates integrate new information gained from the simulated rollout, gradually improving the accuracy of value estimates for different reasoning paths. As a result, subsequent selection steps—guided by the UCB criterion—are better able to balance exploration of less-visited actions with exploitation of those that have consistently yielded high returns.  

Through repeated iterations of simulation and backpropagation, the MCTS framework converges toward identifying reasoning trajectories that are both semantically consistent with the query and structurally grounded in the graph. Over time, this iterative refinement enables the agent to uncover robust, interpretable, and high-quality reasoning paths, thereby improving the reliability and factual correctness of the final answers generated by \myModel.

\subsection{End-to-End Reinforcement Learning}

While MCTS provides a principled mechanism for exploring reasoning trajectories, its effectiveness ultimately depends on how well the explored paths are evaluated and refined. To fully exploit this process, we leverage the information collected during MCTS to  fine tune the reasoning policy $\pi_\theta$ through end-to-end reinforcement learning. 
Specifically, we adopt Group Relative Policy Optimization (GRPO)~\cite{grpo}, an advanced policy optimization algorithm that directly optimizes $\pi_\theta$ with respect to the paths discovered by MCTS. For each trajectory, we introduce an {outcome-directed reward function} $R(\tau)$, which jointly measures the structural validity of the reasoning process and its semantic alignment with the original query. This design ensures that the policy is guided toward producing both faithful and well-structured answers.  

More specifically, given a dataset of questions $q \in \mathcal{D}_Q$, the agent interacts with the text-attribute graph $G$ via MCTS to generate a collection of reasoning trajectories $\{\tau_i\}_{i=1}^N \subseteq \mathcal{T}_q$, where each trajectory is defined as $\tau_i = \big((s^{(i)}_0, a^{(i)}_0), (s^{(i)}_1, a^{(i)}_1), \ldots, (s^{(i)}_T, a^{(i)}_T)\big)$.
Based on the collected trajactory, we utilize GRPO to fine tune the model. The GRPO-based training objective is formulated as:
\begin{align*}
&J_{\text{GRPO}}(\theta) 
= \mathbb{E}_{\![\, s_0 \sim \{P(q)\mid q \in \mathcal{D}_Q\}, \; \{\tau_i\}_{i=1}^N \sim \pi_{\theta_{\text{old}}}(\mathcal{T}_q \mid s_1; G) \,]} \\ 
&\Bigg[\frac{1}{N}\sum_{i=1}^N \frac{1}{|\tau_i|}\sum_{t=1}^{|\tau_i|} 
\min \bigg( \rho_\theta(a_t^{(i)}) \hat{A}(\tau_i), \, 
\text{clip}\big(\rho_\theta(a_t^{(i)}), 1 \pm \epsilon\big)\hat{A}(\tau_i)\bigg) \nonumber \\
&- \beta D_{\text{KL}}(\pi_\theta \parallel \pi_{\text{ref}}) \Bigg], 
\end{align*}
where $\rho_\theta(a_t^{(i)}) = 
\frac{\pi_\theta(a_t^{(i)} \mid s_{t-1}^{(i)}; G_H)}{\pi_\theta^{\text{old}}(a_t^{(i)} \mid s_{t-1}^{(i)}; G_H)}$
and $\hat{A}(\tau_i) = \frac{R(\tau_i) - \text{mean}\big(\{R(\tau_j)\}_{j=1}^N\big)}{\text{F}_{\text{norm}}\big(\{R(\tau_j)\}_{j=1}^N\big)}$.
Here, $\rho_\theta$ is the importance weight that corrects for distributional shift, $\hat{A}(\tau_i)$ denotes the normalized advantage (e.g., using standard deviation as the normalization function), the clipping operator $\text{clip}()$ stabilizes optimization, and the KL penalty encourages $\pi_\theta$ to remain close to a reference policy $\pi_{\text{ref}}$, and  $\beta$ denotes the regularization strength.

\paragraph{Reward Function.} To ensure high-quality reasoning outcomes, we design a trajectory-level reward function $R(\tau)$ that considers both structural coherence and semantic correctness. It combines a \emph{format reward} $R_{\text{format}}(\tau)$ and a \emph{reasoning reward} $R_{\text{reasoning}}(\tau)$, promoting structured multi-step reasoning and accurate answers.

{(i) Format Reward.}  
The format reward encourages the model to produce outputs following the required reasoning structure. At each step $(s_t, a_t)$, the output should contain identifiable reasoning components such as \texttt{Thought}, \texttt{Action}, and \texttt{Observation}. Each correctly formatted element contributes $0.5$ to the reward, capped at $1.0$:
\begin{align*}
R_{\text{format}}(\tau) = \min\Big(1.0, \, 0.5 \cdot \sum_{t=1}^T \mathbb{I}\{(a_t^{\text{think}}, \alpha_t^{\text{Act}}, a_t^{\text{Obs}})
\text{ is well-formed}\}\Big)
\end{align*}
where $\mathbb{I}\{\cdot\}$ is the indicator function.

{(ii) Reasoning Reward.}  
To promote meaningful graph operations, the model is rewarded for using graph-related functions such as \texttt{RetrieveNode}, \texttt{NeighbourCheck}, \texttt{NodeFeature}, and \texttt{NodeDegree}. Explicitly correct answers further increase the reward, while obvious errors or failures are penalized:
\begin{align*}
R_{\text{reasoning}}(\tau) = 
0.5 \cdot \mathbb{I}\{\text{valid operation}\}
+ 1.0 \cdot \mathbb{I}\{a_T^{\text{ans}} = y_q^\ast\} \\
- 0.5 \cdot \mathbb{I}\{a_T^{\text{ans}} \neq y_q^\ast\}.
\end{align*}

{(iii) Overall Reward.}  
The total trajectory reward integrates both format and reasoning components, with semantic correctness contributing only if the reasoning format is valid:
\begin{align}
R(\tau) = \max\big(-1.0, \min\big(3.0, R_{\text{format}}(\tau) + R_{\text{reasoning}}(\tau)\big)\big),
\end{align}
where the reward is clipped to the stable range $[-1.0, 3.0]$ for training stability. This design directly reinforces structurally coherent and semantically accurate multi-step reasoning over the graph $G$, encouraging faithful and well-grounded answers.

%% file: 005experiments.tex
\begin{table}[t]
\centering
\caption{Dataset statistics for GRBENCH, as introduced by ~\cite{jin2024graphchainofthoughtaugmentinglarge}.}
\label{tab:grbench-stats}
\small
\setlength{\tabcolsep}{1.5pt} 
\begin{tabular}{l@{\hskip 3pt}l@{\hskip 3pt}c@{\hskip 3pt}c@{\hskip 3pt}c@{\hskip 3pt}c}
\toprule
\textbf{Domain} & \textbf{Topic} & \textbf{\# Nodes} & \textbf{\# Edges} & \textbf{\# Templates} & \textbf{\# Questions} \\
\midrule
\multirow{6}{*}{Academic} 
& CS                & $\sim$8M   & $\sim$52M  & 15 & 150 \\
& Biology           & $\sim$4M   & $\sim$39M  & 14 & 140 \\
& Chemistry         & $\sim$4M   & $\sim$30M  & 14 & 140 \\
& Material Science  & $\sim$3M   & $\sim$22M  & 14 & 140 \\
& Medicine          & $\sim$6M   & $\sim$30M  & 14 & 140 \\
& Physics           & $\sim$2M   & $\sim$33M  & 14 & 140 \\
\midrule
E-commerce   & Amazon     & $\sim$9M   & $\sim$313M & 20 & 200 \\
Literature   & Goodreads  & $\sim$3M   & $\sim$22M  & 24 & 240 \\
Healthcare   & Disease    & 47K        & $\sim$4M   & 27 & 270 \\
Legal        & Freelaw    & $\sim$84M  & $\sim$114M & 18 & 180 \\
\bottomrule
\end{tabular}
\end{table}

This section presents the experimental setup, main results, and analysis. We address the following research questions (RQs): RQ1: Does \myModel\ outperform existing methods? RQ2: How effective is the agent-based approach compared to other RAG methods? RQ3: How much does the choice of backbone LLM affect performance?

\subsection{Experimental Setup}

\textbf{Baselines.}  We compare \myModel\ with three types of baseline methods:  1) \textbf{Base LLMs}~\cite{touvron2023llama2openfoundation,jiang2024mixtralexperts,gpt2}: These models answer questions using only their own knowledge, without any external data. We use standard prompting, giving simple instructions and letting the model generate an answer.  2) \textbf{Text RAG LLMs}: Here, we treat the graph as a text corpus. A retriever selects relevant passages from the graph, which are then used as extra context to help the LLM answer questions.  3) \textbf{Graph RAG and Reasoning LLMs}~\cite{jin2024graphchainofthoughtaugmentinglarge,graphmcts}: This approach extends graph RAG by treating LLMs as agents that traverse the graph to search for answers. 
For all baselines, we test three LLM backbones: LLaMA-2-13b-chat~\cite{touvron2023llama2openfoundation}, Mixtral-8x7b-Instruct~\cite{jiang2024mixtralexperts}, and GPT-3.5-turbo~\cite{openai2024gpt4technicalreport}.  

\textbf{Dataset.} 
We use a text-attributed graph QA dataset covering five domains: {Academic}, {E-commerce}, {Literature}, {Healthcare}, and {Legal}. Each data sample consists of a question-answer pair. The questions are carefully designed to reflect realistic, domain-specific scenarios that can be answered by navigating the text-attributed graph. Detailed statistics of the dataset are provided in Table~\ref{tab:grbench-stats}.

\begin{table*}[t]
\centering
\caption{Model performance on GRBENCH ~\cite{jin2024graphchainofthoughtaugmentinglarge} comparing standard LLMs, text retrieval augmented LLMs (Text RAG), graph retrieval augmented LLMs (Graph RAG), Graph-COT. Results are reported using ROUGE-L (R-L) and GPT-4 Score (GPT4score). Mixtral-8x7b-Instruct is used as the backbone for Graph-COT~\cite{jin2024graphchainofthoughtaugmentinglarge}, GraphMCTS~\cite{graphmcts} and \myModel.}
\label{tab:grbench-results}
\small
\setlength{\tabcolsep}{1pt} 
\begin{tabular}{lcccccccccc}
\toprule
\textbf{Model} & \multicolumn{2}{c}{\textbf{Academic}} & \multicolumn{2}{c}{\textbf{E-commerce}} & \multicolumn{2}{c}{\textbf{Literature}} & \multicolumn{2}{c}{\textbf{Healthcare}} & \multicolumn{2}{c}{\textbf{Legal}} \\
 & R-L & GPT4score & R-L & GPT4score & R-L & GPT4score & R-L & GPT4score & R-L & GPT4score \\
\midrule
\multicolumn{11}{c}{\textit{Base}} \\ \hline
LLaMA-2-13b-chat & 8.13 & 8.03 & 7.01 & 12.00 & 5.32 & 20.83 & 5.25 & 13.70 & 15.97 & 16.11 \\
Mixtral-8x7b     & 9.02 & 8.14 & 12.54 & 18.00 & 7.50 & 22.50 & 3.88 & 20.00 & 12.74 & 16.11 \\
GPT-3.5-turbo    & 6.05 & 12.80 & 9.18 & 23.50 & 10.43 & 26.67 & 5.83 & 14.44 & 10.51 & 20.00 \\
\midrule
\multicolumn{11}{c}{\textit{Text RAG}} \\ \hline
LLaMA-2-13b-chat & 8.69 & 8.52 & 9.23 & 12.50 & 7.61 & 20.00 & 1.44 & 5.93 & 15.37 & 16.67 \\
Mixtral-8x7b     & 8.44 & 8.02 & 23.14 & 29.50 & 13.35 & 27.92 & 3.22 & 16.67 & 19.69 & 25.00 \\
GPT-3.5-turbo    & 5.83 & 9.91 & 14.06 & 20.00 & 10.04 & 20.83 & 4.57 & 8.52 & 18.14 & 23.89 \\
\midrule
\multicolumn{11}{c}{\textit{Graph RAG}} \\ \hline
LLaMA-2-13b      & 22.01 & 22.97 & 12.48 & 20.00 & 9.25 & 20.00 & 2.97 & 4.81 & 17.98 & 17.22 \\
Mixtral-8x7b     & 27.77 & 31.20 & 32.87 & 37.00 & 20.08 & 33.33 & 8.66 & 15.19 & 23.48 & 25.56 \\
GPT-3.5-turbo    & 18.45 & 26.98 & 17.52 & 28.00 & 14.94 & 24.17 & 8.69 & 14.07 & 18.66 & 22.22 \\
\midrule
\multicolumn{11}{c}{\textit{Agent}} \\ \hline
{GRAPH-COT} &  {27.35} &  {29.43} &  {29.38} &  {32.78} &  {32.44} &  {38.75} &  {23.33} &  {26.02} &  {19.81} &  {22.77} \\
{GRAPH-MCTS} &  {31.42} &  {32.16} &  {33.71} &  {37.22} &  {37.52} &  {45.83} &  {26.79} &  {32.84} &  {26.23} &  {27.78} \\
\textbf{\myModel} & \textbf{32.35} & \textbf{33.82} & \textbf{34.61} & \textbf{38.76} & \textbf{38.42} & \textbf{45.83} & \textbf{29.54} & \textbf{33.70} & \textbf{27.36} & \textbf{32.20} \\

\bottomrule
\end{tabular}
\end{table*}

\textbf{Evaluation Metrics.}  We use both rule-based and model-based metrics. For rule-based evaluation, we use ROUGE-L to measure how similar the predicted answers are to the ground truth. For model-based evaluation, we use GPT-4 to check if answers are correct, reporting the percentage of answers labeled ``correct'' as the GPT-4 Score.

All experiments are conducted on NVIDIA H200 GPUs using Python 3.8 and Huggingface 4.36.2. We use \texttt{Mpnet-v2 7} as the retriever for all baselines and our method, and implement the indexing with FAISS~\cite{johnson2019billion}. 
For \myModel, we adopt Mixtral-8x7b-Instruct as the backbone LLM in the main results.  

\subsection{Overall Performance}

Table~\ref{tab:grbench-results} presents the performance of \myModel\ compared to all baselines across the five domains. We can observe several clear trends:
First, {Base LLMs} perform the worst among all methods. This is expected, as they rely solely on their internal knowledge without accessing any external data, limiting their ability to answer domain-specific or complex questions accurately. 
{Text RAG LLMs} show improved performance, especially in domains like \emph{E-commerce}, where relevant textual information can be effectively retrieved from the external graph. However, even with retrieval, these models still struggle in domains that require deeper reasoning or structured information, such as \emph{Healthcare} and \emph{Legal}.
{Graph RAG LLMs} generally outperform Text RAG, demonstrating that including the 1-hop subgraph around retrieved nodes provides richer structural context. This structured graph information allows the LLM to reason more effectively over relationships and dependencies, which is particularly useful in domains with complex interconnections.
The strongest results come from {agent-based approaches}. {GRAPH-COT} combines the reasoning capabilities of LLMs with graph-structured input, achieving significantly higher ROUGE-L and GPT-4 scores than all non-agent baselines. Graph-MCTS~\cite{graphmcts} shows a better performance advantage. Finally, \myModel\ further improves performance by leveraging a more advanced agent framework alongside graph reasoning. It consistently achieves the highest scores across all domains, showing that integrating structured graph reasoning with an LLM-driven agent is an effective strategy for accurate question answering.
These results highlight two key points: (1) incorporating structured graph information is crucial for challenging domains, and (2) an agent-based reasoning framework can fully exploit this information to deliver more reliable and accurate answers.



\subsection{RAG vs LLM as Agent}
To assess the effectiveness of graph retrieval-augmented LLMs, we experimented with different retrieval scopes: a single node, 1-hop ego-graphs, and 2-hop ego-graphs. In each case, the retrieved subgraph is linearized into a text sequence and used as context for the LLM. Table~\ref{RAG} reports the average performance across all datasets. Interestingly, the 1-hop ego-graph setting yields the best results among the three, yet still falls short of the performance achieved by \myModel. A key limitation lies in the exponential growth of nodes and text as the hop count increases. While larger subgraphs do contain more information, they also generate longer contexts, often exceeding the input limits of LLMs and leading to degraded performance. In contrast, \myModel\ offers a more efficient approach by selectively extracting and encoding the most relevant graph information within the model’s capacity.

\begin{table}[t]
\centering
\caption{Results of LLM with different retrieval augmentation methods on GRBENCH.}
\label{RAG}
\begin{tabular}{lc}
\toprule
\textbf{Model} & \textbf{GPT4score} \\
\midrule
GPT-3.5-turbo & 19.48 \\
+ node retrieval & 16.63 \\
+ 1-hop subgraph retrieval & 23.09 \\
+ 2-hop subgraph retrieval & 22.12 \\
+ \myModel & {36.86} \\
\bottomrule
\end{tabular}
	\vspace{-1\baselineskip}
\end{table}

\subsection{Ablation Study}

In the main results, we adopt Mixtral-8x7b-Instruct as the LLM backbone for \myModel. In this part, we explore \myModel\ with different LLM backbones, including LLaMA-2-7b-chat, LLaMA-2-13b-chat, Mixtral-8x7b-Instruct. We randomly extract a subset from GRBENCH for experimentation. The results are shown in Table~\ref{tab:llm-backbones}. 

From the results, we find that the choice of LLM backbone matters. LLMs with stronger instruction-following and reasoning abilities lead to better performance in \myModel.

\begin{table}[h]
\centering
\caption{Results of \myModel\ with different LLM backbones.}
\vspace{-1\baselineskip}
\label{tab:llm-backbones}
\begin{tabular}{lc}
\toprule
Model & GPT4 Score \\
\midrule
LLaMA-2-7b-chat & 7.64 \\
LLaMA-2-13b-chat & 14.58 \\
Mixtral-8x7b-Instruct & 32.33 \\
\bottomrule
\end{tabular}
\end{table}

\begin{table}[t]
\centering
\caption{Effect of search depth and width on reasoning performance in the Healthcare dataset.}
\vspace{-1\baselineskip}
\label{RAG12}
\setlength{\tabcolsep}{1.5pt}
\begin{tabular}{lccc}
\toprule
\textbf{Max Depth} & \textbf{R-L (width=2)} & \textbf{R-L (width=3)} & \textbf{R-L (width=4)} \\
\midrule
6  & 16.42 & 17.35 & 17.62 \\
8  & 26.71 & 27.58 & 27.94 \\
10 & 28.83 & {29.54} & 29.67 \\
12 & 28.96 & {29.85} & 39.91 \\
\bottomrule
\end{tabular}
\vspace{-1\baselineskip}
\end{table}

\noindent\textbf{Ablation Study on MCTS Parameters.} To examine how search settings influence reasoning performance, we analyze two key MCTS parameters: the maximum depth (\texttt{max\_depth}) and the search width (\texttt{width}). As shown in Table~\ref{RAG12}, shallow searches (e.g., depth = 6) often stop too early, leading to incomplete reasoning and lower scores. Increasing the depth to 8–10 improves multi-hop reasoning, while deeper searches ($>12$) add little benefit but increase cost.  
Wider searches explore more candidate paths, improving accuracy slightly but requiring much more computation. A moderate width of 2–3 achieves a good balance between performance and efficiency.

\noindent\textbf{GRPO Effecitiveness.}
We investigate the impact of the GRPO optimization on \myModel's reasoning performance by comparing results with and without GRPO (Table~\ref{tab:grbench-results-rl}). Removing GRPO leads to drops in ROUGE-L scores. This indicates that the end-to-end reinforcement learning component plays a important role in guiding the LLM agent to select more effective graph traversal and reasoning actions, thereby improving multi-hop question answering performance.

\begin{table}[t]
\centering
\caption{Model performance on GRBENCH using ROUGE-L.}
\vspace{-1\baselineskip}
\label{tab:grbench-results-rl}
\small
\setlength{\tabcolsep}{1.0pt} 
\begin{tabular}{lccccc}
\toprule
\textbf{Model} & \textbf{Academic} & \textbf{E-commerce} & \textbf{Literature} & \textbf{Healthcare} & \textbf{Legal} \\
\midrule
Without GRPO & 31.42	 & 33.71	 & 37.52	 & 26.79	 & 26.23 \\
With GRPO    & \textbf{32.35} & \textbf{34.61} & \textbf{38.42} & \textbf{29.54} & \textbf{27.36} \\
\bottomrule
\end{tabular}
\end{table}

%% file: 006related_work.tex
\paragraph{LLM for Text Attributed Graph Question Answering} Graph-based question answering has been studied for many years. Recently, Large Language Models (LLMs) have sparked new interest in reasoning over text-attributed graphs. For example, GenTKG~\cite{liao2023gentkg} shows that using LLaMA2-7B as the backbone improves cross-domain generalization and reduces the need for large training datasets in temporal tasks. KG-GPT~\cite{kim2023kg} combines LLMs with graph retrieval to help the model identify relevant relations. CSProm-KG~\cite{chen2023dipping} uses prompt tuning to integrate both structural and textual features, while ECOLA~\cite{han2022ecola} leverages LLMs to complete graph data and jointly optimize knowledge-text prediction.
Although these methods are promising, they often face scalability issues—large models are expensive to retrain and fine-tune as data and task complexity increase. Retrieval-Augmented Generation (RAG) has emerged as a practical alternative~\cite{replug}. Instead of retraining, RAG brings in external knowledge at inference time, which reduces hallucinations and improves factual accuracy. The simplest version, naive RAG~\cite{edge2025localglobalgraphrag}, follows a basic retrieve-and-generate workflow. More advanced methods improve retrieval with query rewriting or expansion~\cite{qa_rewrite}, and further refine results through reranking~\cite{yin2024rethinking}.
In parallel, LLMs are being used in agent-based frameworks. Chain-of-Thought prompting~\cite{jin2024graphchainofthoughtaugmentinglarge} encourages step-by-step reasoning, while ReAct~\cite{react} combines reasoning with tool use. These ideas have inspired further work on multi-agent collaboration~\cite{guo2024largelanguagemodelbased} and stronger planning abilities.

\paragraph{Reinforcement learning for LLMs Optimization}
Reinforcement learning has emerged as a powerful paradigm for optimizing large language models (LLMs), and several recent approaches extend beyond traditional human-in-the-loop methods such as PPO or DPO. For example, MCTS-DPO~\cite{MCTS_DPO} integrates Monte Carlo Tree Search into DPO training by generating multiple candidate outputs and treating sibling nodes as comparative training examples, with high-Q-value nodes serving as positive samples to guide fine-tuning. Iterative DPO~\cite{srlm} further advances this idea by introducing self-rewarding language models, where the model generates and evaluates its own outputs, thereby enhancing instruction following and self-assessment without requiring external feedback. Reinforced Self-Training (ReST)~\cite{rest} adopts a grow-and-improve strategy, alternating between generating candidate outputs and fine-tuning only on those surpassing a gradually increasing quality threshold, which ensures continuous improvement. Beyond optimizing final outputs, reinforcement learning has also been applied to process-level supervision, where intermediate reasoning steps are annotated and rewarded. Lightman et al.~\cite{lightman2023letsverifystepstep} show the effectiveness of process reward models (PRMs) with PRM800K, while automated alternatives such as stronger external LLM annotation~\cite{luo2025wizardmathempoweringmathematicalreasoning}, Monte Carlo simulation~\cite{wang2024mathshepherdverifyreinforcellms}, and Monte Carlo Tree Search (MCTS)–based evaluation~\cite{luo2024improvemathematicalreasoninglanguage} reduce reliance on human labeling. Finally, Zhang et al.~\cite{zhang2023largelanguagemodelssemiparametric} propose a self-refining loop that iteratively improves annotations via PRM-guided MCTS, enabling reinforcement learning at both output and process levels. Together, these approaches demonstrate the growing role of reinforcement learning in advancing the optimization of LLMs across diverse reasoning tasks.

%% file: 007conclusion.tex
This paper introduces \myModel, a framework that enables Large Language Models (LLMs) to perform structured, step-by-step reasoning over text-attributed graphs. By integrating Monte Carlo Tree Search (MCTS) with the generative capabilities of LLMs, \myModel\ selectively and iteratively explores the graph, focusing only on the most relevant nodes and textual information. This approach improves both reasoning depth and computational efficiency. To further enhance performance, we employ reinforcement learning with Grouped Reinforcement Policy Optimization (GRPO), which optimizes the model’s decision-making during graph traversal. Experimental results demonstrate that \myModel\ achieves superior performance compared to all baseline methods.